\documentclass[10pt,twocolumn,letterpaper]{article}

\usepackage{iccv}              % To produce the CAMERA-READY version
\definecolor{iccvblue}{rgb}{0.21,0.49,0.74}
\usepackage[pagebackref,breaklinks,colorlinks,allcolors=iccvblue]{hyperref}
\usepackage{algorithm}
\usepackage{algpseudocode}
\usepackage{soul}
\usepackage{multirow}
\usepackage{pifont}
\usepackage{colortbl}
\usepackage{caption}
\usepackage{float}
\usepackage{threeparttable}
\usepackage[accsupp]{axessibility}  % Improves PDF readability for those with disabilities

\def\paperID{9134} % *** Enter the Paper ID here
\def\confName{ICCV}
\def\confYear{2025}

\title{Versatile Transition Generation with Image-to-Video Diffusion}

\author{
    Zuhao Yang$^{1}$\footnotemark[1]\quad
    Jiahui Zhang$^{1}$\quad
    Yingchen Yu$^{2}$\quad
    Shijian Lu$^{1}$\footnotemark[2]\quad
    Song Bai$^{2}$ \\ [0.5em]
    $^{1}$Nanyang Technological University\qquad
    $^{2}$ByteDance Inc. \\ [0.5em] \href{https://mwxely.github.io/projects/yang2025vtg/index}{\textcolor{NavyBlue}{https://mwxely.github.io/projects/yang2025vtg/index}}
}

\begin{document}
\twocolumn[{
    \renewcommand\twocolumn[1][]{#1}
    \maketitle
    \begin{center}
        \centering
        \captionsetup{type=figure}
        \includegraphics[width=\textwidth]{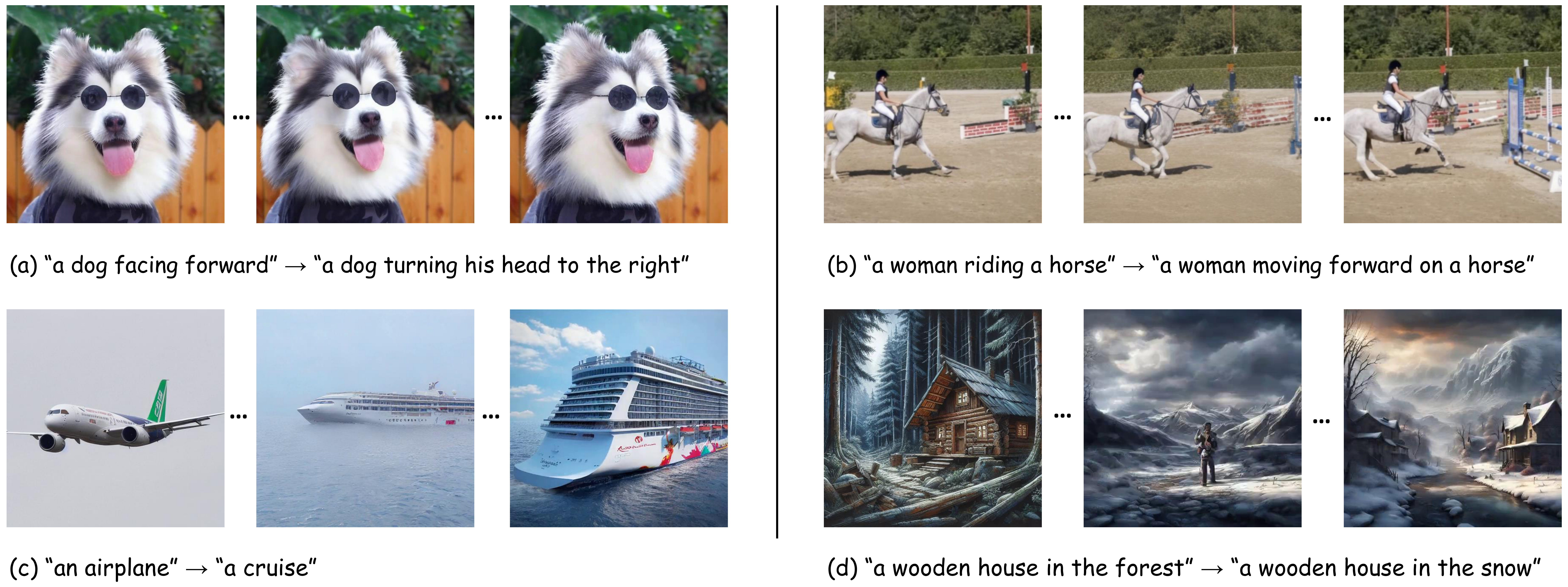}
        \captionof{figure}{Illustration of \emph{Versatile Transition Generation} (VTG). VTG is capable of performing four types of transition generation, namely, (a) object morphing, (b) motion prediction, (c) concept blending, and (d) scene transition, within a single uniform framework.
        }
        \label{fig:teaser}
    \end{center}
}]
{\renewcommand*\thefootnote{\fnsymbol{footnote}}
    \footnotetext[1]{This work was done while Zuhao Yang was interning at ByteDance.}
    \footnotetext[2]{Shijian Lu is the corresponding author.}
}
\begin{abstract}
Leveraging text, images, structure maps, or motion trajectories as conditional guidance, diffusion models have achieved great success in automated and high-quality video generation. However, generating smooth and rational transition videos given the first and last video frames as well as descriptive text prompts is far underexplored. We present \textbf{VTG}, a \textbf{\underline{V}}ersatile \textbf{\underline{T}}ransition video \textbf{\underline{G}}eneration framework that can generate smooth, high-fidelity, and semantic-coherent video transitions.
VTG introduces interpolation-based initialization that helps preserve object identity and handle abrupt content changes effectively. In addition, it incorporates dual-directional motion fine-tuning and representation alignment regularization that mitigate the limitations of the pre-trained image-to-video diffusion models in motion smoothness and generation fidelity, respectively.
To evaluate VTG and facilitate future studies on unified transition generation, we collected \textbf{TransitBench}, a comprehensive benchmark for transition generation that covers two representative transition tasks including concept blending and scene transition. Extensive experiments show that VTG achieves superior transition performance consistently across the four tasks.
\end{abstract}
    
\section{Introduction}
\label{sec:intro}

The success of diffusion models \cite{ho2020denoising, song2020score} in image synthesis \cite{nichol2021glide, saharia2022photorealistic, ramesh2022hierarchical, rombach2022high} has inspired a number of studies on diffusion-based video synthesis \cite{ho2022imagen, zhou2022magicvideo, wang2023lavie, zhang2023i2vgen}. Leveraging textual prompts, video frames, structure maps and even motion patterns, several studies \cite{singer2022make, chen2024videocrafter2, tu2023motioneditor, wang2023motionctrl} have demonstrated impressive synthesis performance by generating realistic and high-fidelity videos automatically. Nevertheless, despite its significant value in various real-world tasks such as video and film production, generating high-quality transition videos conditioned on the first and last frames together with related text prompts remains largely underexplored.

Generating realistic transition videos is a non-trivial task. A high-quality transition generator should meet at least four criteria: 1) semantic similarity to the input frames; 2) high fidelity to the input frames; 3) smoothness across the generated frames; and 4) alignment with the provided text prompts. Moreover, the research community in transition generation mostly relies on self-collected and well-filtered videos that are not accessible to the public~\cite{chen2023seine, zeng2023make}. This further hinders the advancement of the research in transition video generation.

Most existing studies tackle the challenge of transition generation via two representative approaches. The first focuses on morphing, given two images of topologically similar objects. Most recent methods~\cite{yang2023impus, zhang2023diffmorpher} harness various deep interpolation techniques for generating rational object-level transition. However, they generate intermittent images rather than temporally coherent video frames, losing the transition smoothness especially while handling moving objects.
The second focuses on video frame interpolation. Most existing studies attempt to estimate intermediate optical flows \cite{reda2022film, li2023amt, zhang2023extracting, liu2024sparse}, or leverage frame conditioning at training time \cite{blattmann2023align, blattmann2023stable, xing2023dynamicrafter, jain2024video}. However, they tend to generate irrational object transitions with abrupt content changes, struggle with producing long transition sequences, and require time-consuming training on large-scale motion videos.
On top of the above, existing studies either employ deep interpolation for conceptual blending transitions \cite{yang2023real, he2024aid} or incorporate a random-mask condition layer for scene transition \cite{chen2023seine}, lacking a uniform framework that can work for multiple transition generation tasks.
% This prompts a compelling inquiry: \emph{can we design a versatile transition generator that works for various transition tasks without extra training on task-specific data?}
This prompts a compelling inquiry: \emph{can we design a versatile transition generator that can handle various transition tasks with minimal adaptation across tasks?}

As illustrated in \Cref{fig:teaser}, we unify four transition tasks to establish \emph{versatile transition generation}, based on the types of input frames:
(1) Object Morphing: The input frames are either the same object with different postures or different objects, as long as they are topologically similar; 
(2) Concept Blending: The input frames contain conceptually different objects (e.g., `an airplane' and `a cruise');
(3) Motion Prediction: The input frames can be seen as two moments in a video containing one (or more) moving object;
(4) Scene Transition: The input frames are conceptually related scene images yet they either belong to two different domains (e.g., `a wooden house in the forest' and `a wooden house in the snow') or represent two distinct components of a scene (e.g., `erupting volcano' and `hot lava').

We design VTG, a versatile transition generation framework built upon image-to-video diffusion models. VTG features three designs for generating smooth, high-fidelity, and semantic-coherent transitions.
First, we employ interpolation-based initialization that effectively mitigates abrupt content changes while handling input frames with significantly different contents. Specifically, we spherically interpolate latent Gaussian noises of the two input frames and text embeddings of corresponding transition captions and leverage two LoRA-integrated U-Nets~\cite{hu2021lora} to capture object-level semantics in the denoising steps.
Second, we adopt a lightweight fine-tuning strategy that merges the predicted forward and backward noises which greatly improves the transition smoothness in motion predictions. The bidirectional sampling can be achieved by fine-tuning the pretrained U-Net with a small collection of videos.
Third, we introduce regularization with self-supervised visual encoder for video diffusion models to explicitly induce feature alignment, which helps learn meaningful representations and enhance the fidelity of generated transition videos.
These components are logically cohesive and, in combination, distinguish VTG from prior work by enabling a single framework to handle diverse transition tasks with minimal task-specific adjustments.
In addition, we collected \emph{TransitBench}, a new benchmark of 200 pairs with first and last frames for concept blending and scene transition.

The contributions of this work can be summarized in three aspects.
First, we propose a unified task of \emph{versatile transition generation}, aiming at smooth and rational transition for object morphing, concept blending, motion prediction, and scene transition.
Second, we design VTG, a novel and versatile framework that can generate semantic-relevant, high-fidelity and temporally coherent video transitions effectively.
Third, we introduce TransitBench, a curated dataset for benchmarking concept blending and scene transition. With TransitBench and other public benchmarks, we demonstrate, both qualitatively and quantitatively, that VTG outperforms the state-of-the-art consistently across the four transition generation tasks.

\section{Related Work}
\label{sec:formatting}

\textbf{Image Morphing.} Image morphing has been a long-standing problem in the computer graphics community \cite{wolberg1998image, zope2017survey, aloraibi2023image}. 
Conventional morphing techniques either utilize correspondence-driven bidirectional image warping \cite{beier1992feature, bhatt2011comparative, darabi2012image, liao2014automating} or minimize the path energy on a Riemannian manifold \cite{miller2001group, shechtman2010regenerative, fish2020image, rajkovic2023geodesics} to obtain image transitions.
While producing seamless morphing, they either require massive human involvement or fail to create new content beyond the two given images.
Recently, several studies~\cite{yang2023impus, zhang2023diffmorpher} explored the prior knowledge in image diffusion models for real-world morphing with more object categories.
Differently, we exploit a pretrained video diffusion model for image morphing, naturally enabling coherent transitions across frames. \\

\noindent\textbf{Video Frame Interpolation.} Video frame interpolation (VFI) has been extensively studied in the computer vision research community \cite{dong2023video}.
Classic VFI algorithms are closely related to optical flow prediction, which obtain interpolation by either forward warping \cite{niklaus2020softmax} or backward warping \cite{kong2022ifrnet, huang2022real, li2023amt} based on the estimated flows.
Several follow-up studies \cite{reda2022film, liu2024sparse, jain2024video} focus on large motion problem in VFI.
Recently, a line of research \cite{feng2024explorative, wang2024generative} explores bounded video generation with image-to-video diffusion models under the guidance of the start and end frames. This approach fuses outputs from forward and backward paths by interpolating the predicted forward noise and its reversed counterpart.
While generating coherent motion, it overlooks the identity preservation and struggles to produce natural transitions when the two input frames differ significantly in content. \\

\noindent\textbf{Transition Generation.} Transition generation is a far underexplored problem, largely due to its open-ended nature.
Typically, a transition generation framework leverages two input frames (i.e., the first and last frames) and a transition caption to produce a scene-level transition video that connects two distinct narrative moments.
This has been explored in~\cite{chen2023seine} that incorporates random masking to selectively suppress information from the original latent code, allowing capturing transition subtleties between frames.
In addition, \cite{xing2023dynamicrafter, zeng2023make} inject the first and last frames as additional conditions by concatenating them with noisy latent codes along the channel dimension.
Recently, a concurrent study \cite{zhang2024tvg} explores transition video generation with image-to-video diffusion as well.
Differently, we formulate the task by \emph{versatile transition generation} by unifying four transition tasks under the same framework, which greatly expands the applicability and significantly enhances the quality of the generated transition videos.

\section{Method}
\label{sec:method}

\begin{figure*}
    \centering
    \includegraphics[width=0.87\linewidth]{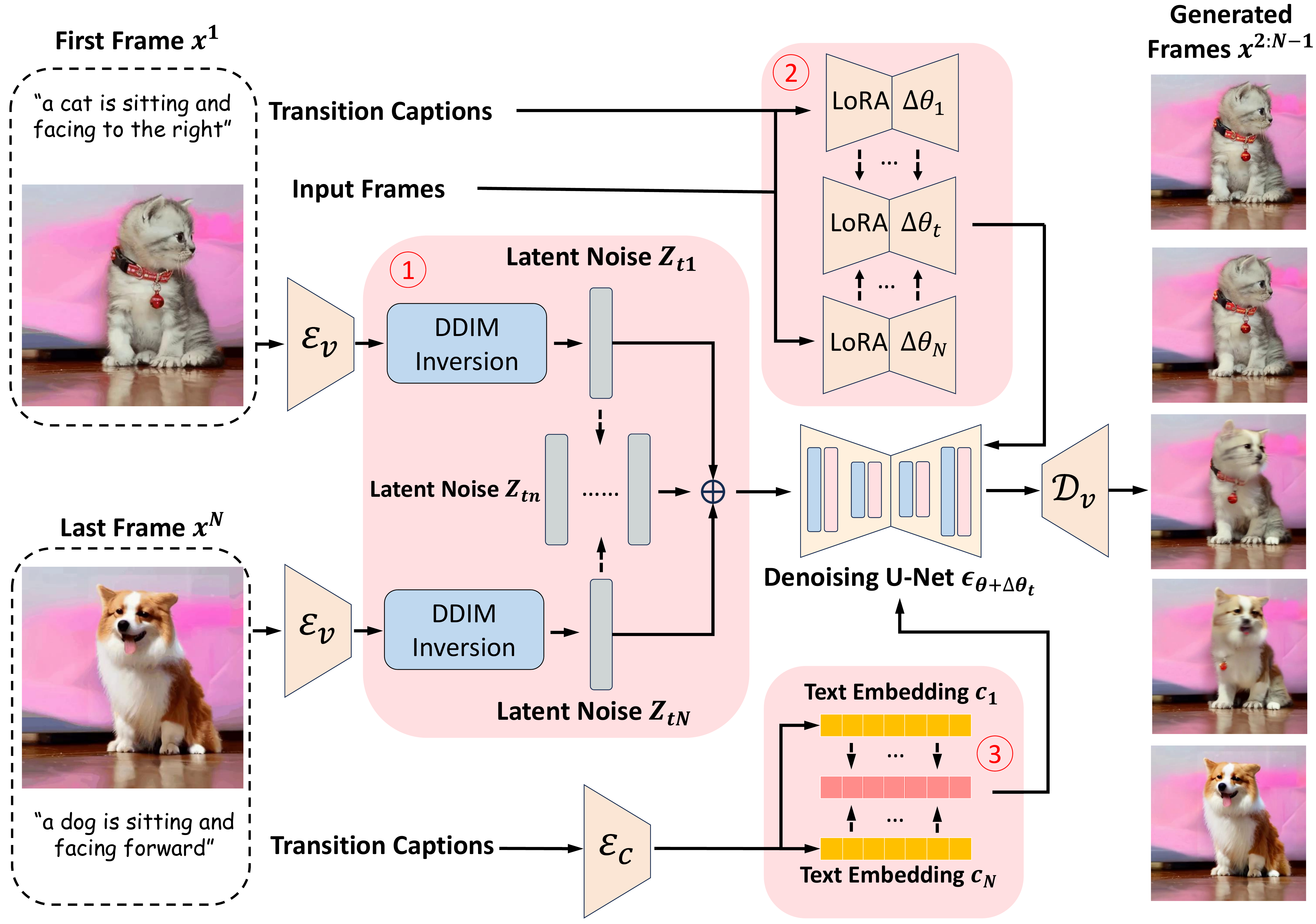}
    \caption{\textbf{Inference framework of VTG.} Our interpolation-based initialization features three designs: \textcolor{red}{\ding{172}} Interpolated Noise Injection, \textcolor{red}{\ding{173}} LoRA Interpolation, and \textcolor{red}{\ding{174}} Frame-aware Text Interpolation. VTG first converts two encoded input frames to latent noises via DDIM inversion. Then, it interpolates between the two latent noises and concatenate intermediate noises along the temporal dimension. To capture meaningful semantics and enable the transition between two conceptually different objects, we employ LoRA interpolation and text interpolation, respectively.}
    \label{fig:framework_infer}
\end{figure*}

\subsection{Preliminaries}
\label{sec:pre}

Latent diffusion models \cite{rombach2022high} are text-to-image diffusion models that operate in a compressed latent space.
During forward diffusion process, it encodes an image sample $\mathbf{x}_0$ into a latent code $\mathbf{z}_0$ and corrupts $\mathbf{z}_0$ by Gaussian noise at each time step $t = 1, \dots, T$:
\begin{equation}
    q(\mathbf{z}_{t}|\mathbf{z}_{t-1}) = \mathcal{N}(\mathbf{z}_{t};\sqrt{1-\beta_{t}}\mathbf{z}_{t-1}, \beta_{t}\mathbf{I}),
\end{equation}
where $\beta_t$ denotes the noise strength at time step $t$ and $\mathbf{I}$ denotes the identity matrix with the same dimensions as $\mathbf{z}_0$.
Given a text embedding $\mathbf{c}$, the backward denoising process can be parameterized as:
\begin{equation}
    p_{\theta}(\mathbf{z}_{t-1}|\mathbf{z}_{t}) = \mathcal{N}(\mathbf{z}_{t-1};\mu_{\theta}(\mathbf{z}_{t},\mathbf{c},t), {\Sigma_{\theta}(\mathbf{z}_{t},\mathbf{c},t)}),
\end{equation}
where $\mu_{\theta}(\mathbf{z}_{t},\mathbf{c},t)$ is output by a U-Net \cite{ronneberger2015u}, and ${\Sigma_{\theta}(\mathbf{z}_{t},\mathbf{c},t)}$ is determined by a noise scheduler (e.g., DDPM \cite{ho2020denoising} or DDIM \cite{song2020denoising}).
The model learns to predict noise via a parameterized noise estimator $\epsilon_\theta$ under the guidance of an objective function:
\begin{equation}
    \mathcal{L}_t = \mathbb{E}_{t\sim[1,T],\mathbf{z}_{0},\mathbf{c},\epsilon_t\sim\mathcal{N}(\mathbf{0},\mathbf{I})}\parallel \epsilon_t-\epsilon_\theta(\mathbf{z}_t,\mathbf{c},t)\parallel^2_2.
\end{equation}

Our framework is built upon pretrained image-to-video diffusion models that incorporate temporal convolutions and temporal attention layers on top of latent diffusion models to establish temporal correlations between video frames.
Given a video $\mathbf{x}_{0} \in \mathbb{R}^{N \times 3 \times H \times W}$, where $N$ is the number of frames and $H \times W$ indicates the image resolution, each frame is first encoded into a latent representation $\mathbf{z}_{0} \in \mathbb{R}^{N \times C \times h \times w}$ via the VAE encoder $\mathcal{E}_{v}$. Both forward diffusion process and backward denoising process are then performed in the latent space. The generated video frames can be collectively obtained through the decoder $\mathcal{D}_{v}$.
To inject image conditions, we fuse and incorporate text-conditioned and image-conditioned features by projecting the conditional image into a text-aligned embedding space. Meanwhile, the conditional image is concatenated with the initial per-frame noise to preserve more visual details.

\subsection{Interpolation-based Initialization}
\label{sec:interp}

% noise interpolation
In diffusion models, the initial Gaussian noise determines the coarse structure that emerges in the early denoising steps, while high-frequency details are refined later.
Since existing image-to-video diffusion models \cite{chen2023seine, xing2023dynamicrafter} randomly initialize the latent code $\mathbf{z}_{0}$ at inference time, each intermediate frame follows a distinct stochastic trajectory: colors, micro-textures, or even object pose shift slightly from frame to frame, yielding perceptual ``flicker'' \cite{deng2024infinite}.

To suppress this drift, we correlate the frame-wise latents by interpolating between the endpoint noises instead of resampling them independently.
Prior work \cite{bodin2025linear} indicates that linear interpolation often yields intermediate latent norms that are extremely unlikely under a unit Gaussian, whereas spherical linear interpolation (SLERP) preserves the Euclidean norm and keeps samples on-distribution. 
Therefore, we leverage SLERP between the latent noises of two input frames $\mathbf{z}_{t1}$ and $\mathbf{z}_{tN}$ as follows: 
\begin{equation}
    \mathbf{z}_{tn} =\frac{\sin ((1-\lambda_{noise}) \phi)}{\sin \phi} \mathbf{z}_{t1} + \frac{\sin (\lambda_{noise} \phi)}{\sin \phi} \mathbf{z}_{tN},
\end{equation}
where $\phi = \arccos \left(\frac{\mathbf{z}_{t1}^T \cdot \mathbf{z}_{tN}}{\left\|\mathbf{z}_{t1}\right\|\left\|\mathbf{z}_{tN}\right\|}\right)$, and $\lambda_{\text{noise}} \in [0, 1]$ denotes the parameter for latent interpolation.
As shown in \Cref{fig:framework_infer}, we concatenate $\mathbf{z}_{tn}$ with $\mathbf{z}_{t1}$ and $\mathbf{z}_{tN}$ along the frame dimension and inject them as the initial latent noise for DDIM sampling.
It is worth noting that we only inject the interpolated latent noises at early denoising steps to preserve the appearance and motion priors from the image-to-video backbone as much as possible.
With such meticulously designed latent injection strategy, our approach addresses the issue of random and abrupt transition frames.

% lora interpolation
Unlike GANs \cite{goodfellow2014generative}, whose latent space is structured and semantically meaningful, diffusion models operate in an unstructured noise space that does not explicitly encode high-level semantics \cite{zhang2023diffmorpher}.
This lack of structure can lead to artifacts when interpolating between semantically different inputs.
To suppress such transition artifacts, we first train two LoRAs $\Delta\theta_1$, $\Delta\theta_N$ given two input frames $x^1$ and $x^N$, optimizing the following objective:
\newcommand{\xt}{\sqrt{\bar\alpha_t}\,\mathbf{z}_{0n}+\sqrt{1-\bar\alpha_t}\,\epsilon}
\begin{equation}
    \resizebox{0.89\linewidth}{!}{$
    \mathcal{L}(\Delta\theta_t)=
    \mathbb{E}_{\epsilon,t}\!\bigl[
      \lVert\,\epsilon-
      \epsilon_{\theta+\Delta\theta_i}\!\bigl(\xt,t,\mathbf{c}_n\bigr)
      \rVert^{2}\bigr]
    $},
\end{equation}
where $\mathbf{z}_{0n}$ is the encoded latent vector of frame $n$, and $\mathbf{c}_n$ is the text embedding associated with transition caption $c^n$.
Then, we linearly interpolate the two LoRAs to fuse the semantics of two input frames:
\begin{equation}
    \Delta\theta = (1-\lambda_{LoRA})\Delta\theta_1 + \lambda_{LoRA}\Delta\theta_N,
\end{equation}
where $\lambda_{LoRA}$ denotes the parameter during LoRA interpolation.

% text interpolation
Typically, video diffusion models \cite{chen2023seine, xing2023dynamicrafter} employ only one caption as the text condition. This hinders the transition generation between two conceptually different objects, causing abrupt content changes without generating intermediate frames with hybrid meanings.
However, for image-to-video diffusion models, the text embedding of the entire frame sequence $c$ is integrated within the denoising U-Net via the cross-attention layer, without explicitly defining per-frame text embedding.
A related \emph{still-image} approach \cite{yang2023impus} linearly blends the two text embeddings to morph between two static endpoints, whereas our method performs \emph{frame-aware} SLERP across prompt sequences as follows:
\begin{equation}
    \mathbf{c}_{\lambda_{text}} =\frac{\sin ((1-\lambda_{text}) \phi)}{\sin \phi} \mathbf{c}_{1} + \frac{\sin (\lambda_{text} \phi)}{\sin \phi} \mathbf{c}_{N},
\end{equation}
where $\lambda_{text} \in [0, 1]$ serves as the frame-aware coefficient to control the transition sequence; $\mathbf{c}_{1}$ and $\mathbf{c}_{N}$ denote the text embeddings of the first and last frames, respectively.

\subsection{Bidirectional Motion Prediction}
\label{sec:dual}

\begin{figure*}
    \centering
    \includegraphics[width=0.89\linewidth]{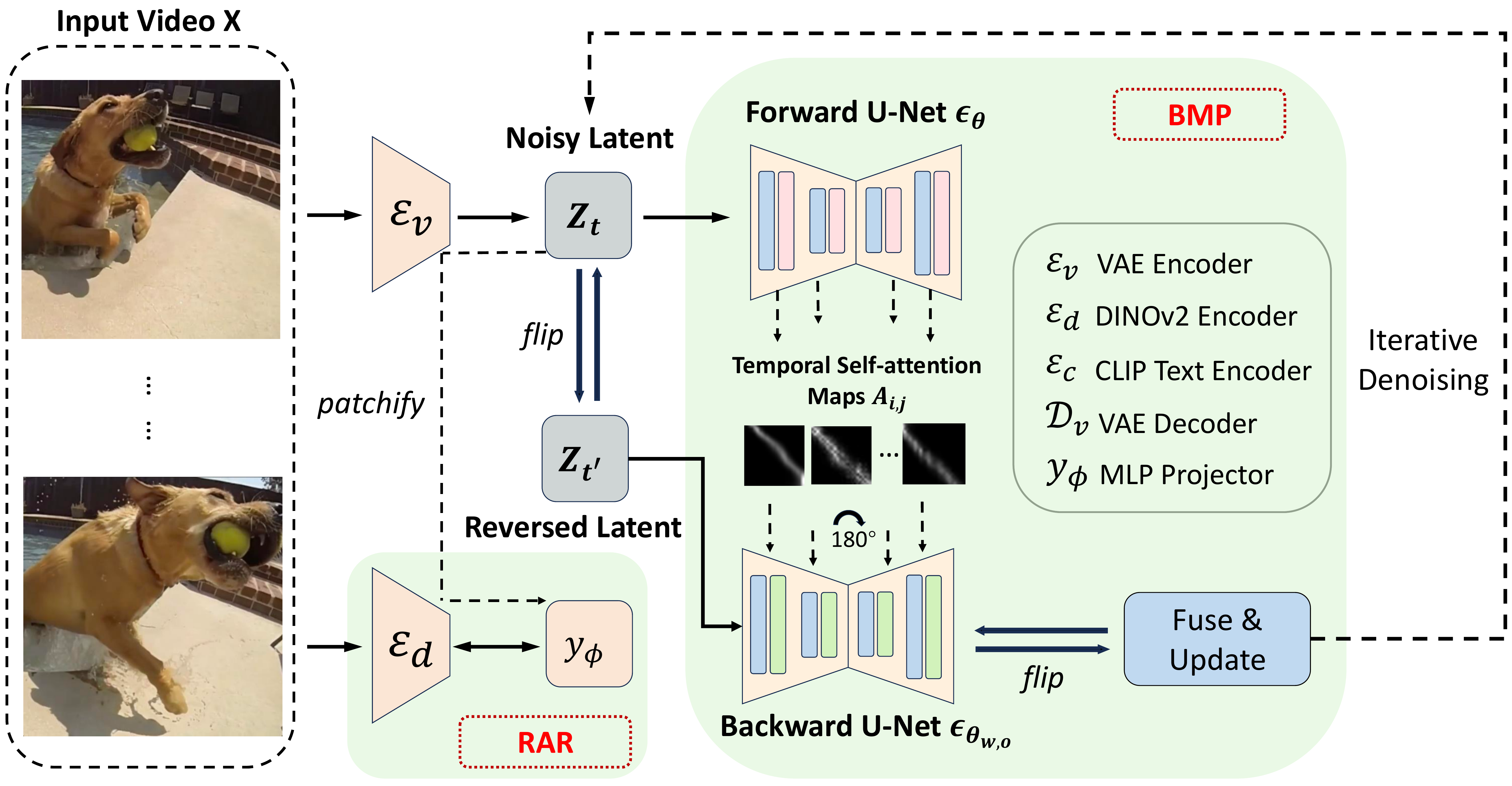}
    \caption{\textbf{Training framework of VTG.} In Bidirectional Motion Prediction (\textcolor{red}{BMP}), the noisy latent is flipped along temporal dimension, and self-attention maps undergo a 180-degree rotation to establish reversed motion-time correlations. Two U-Nets separately predict forward and backward motion, with the backward noise reversed again to fuse with the forward noise, ensuring a consistent motion path for iterative denoising. In Representation Alignment Regularization (\textcolor{red}{RAR}), we spatially patchify each frame independently, then aggregate the per-patch alignment loss across temporal dimension.}
    \label{fig:framework_train}
\end{figure*}

In our experiments, we observed notable quality differences when reversing the order of input frames.
This discrepancy arises from two main factors: (1) the model shows a bias toward resembling the initial input frame more closely due to conditional image leakage \cite{zhao2024identifying}; (2) existing image-to-video diffusion models \cite{chen2023seine, xing2023dynamicrafter} are pretrained exclusively for forward motion prediction, while real-world motion is naturally asymmetric, resulting in ambiguity when predicting reverse motion.

Inspired by \cite{wang2024generative}, we simultaneously predict forward and backward motions to mitigate the ambiguity issue. Specifically, we rotate the temporal self-attention maps by 180 degrees both horizontally and vertically, causing a reversal of attention relationships as illustrated in \Cref{fig:framework_train}.
We also reverse the original forward trajectory of video \{$x^1,x^2,\dots,x^{N}$\} to \{$x^{N},x^{N-1},\dots,x^{1}$\}, given the two input frames $x^1$ and $x^{N}$.
Then, we obtain the reversed noisy latent $\mathbf{z}_{t^\prime}$ by flipping $\mathbf{z}_{t}$ along the temporal dimension. After that, $\mathbf{z}_{t^\prime}$ is fed into the 3D U-Net for backward motion prediction.

During training, we employ a lightweight fine-tuning strategy that only updates the parameters of value and output matrices (i.e., $\theta_{w, o}$) in the temporal attention layers. The objective $\mathcal{L}_{BMP}$ is formulated by taking the L2 norm between the predicted noise and ground-truth reversed noise:
\begin{equation}
    \mathcal{L}_{BMP} = \parallel \text{flip}(\epsilon_t)-\epsilon_{\theta_{w,o}}(\mathbf{z}_{t^\prime},\mathbf{c},t,A^\prime_{i,j})\parallel^2_2.
\end{equation}
The predicted backward noise is reversed again to fuse with forward noise. Specifically, we simply employ linear interpolation to ensure forward-backward consistency:
\begin{equation}
    \epsilon_t = (1-\lambda_{BMP})\epsilon_{t,i} + \lambda_{BMP}\epsilon^\prime_{t,N-i},
\end{equation}
where $\lambda_{BMP}=0.5$ denotes the weighting factor.

\subsection{Representation Alignment Regularization}
\label{sec:loss}

Though previous techniques effectively mitigate abrupt content changes and motion-reversal ambiguity, our approach remains susceptible to generating blurry and low-fidelity transitions, especially when input frames contain fine-grained textures (e.g., bicycle spokes or fabric weave).
Recent work \cite{yu2024representation} shows that diffusion latents inherently lack high-frequency semantics compared to self-supervised representations like DINOv2's \cite{oquab2023dinov2}.
Such mismatch tends to accumulate over video frames and appears as blur.

To overcome this, we propose to distill DINOv2 features back into the denoising trajectory of video diffusion.
As illustrated in \Cref{fig:framework_train}, the video latent is first patchified into $N$ sequences of $T$ tokens along the temporal dimension. For a N-frame video, the latent vector $\mathbf{z}_n$ of $n$-th frame has the shape of $C \times h \times w$. Given a pre-defined patch size $s$, the sequence length $T$ should be: $T = (h/s)\times (w/s)$.
Then, we project the per-frame latent representation $\mathbf{z}_n$ of video diffusion via a multilayer perceptron (MLP) to align with DINOv2 representation $\mathbf{y}_\ast = \mathcal{E}_{d}(\mathbf{x}_\ast) \in \mathbb{R}^{P \times m}$, where $\mathbf{y}_\ast$ denotes the encoder output given a clean video frame $\mathbf{x}_\ast$; $P$ and $m$ respectively denotes the number of patches and  the embedding dimension of DINOv2 encoder $\mathcal{E}_{d}$.
Let $y_\phi(\mathbf{h}_t)$ be the projected representation of intermediate diffusion latent feature $\mathbf{h}_t$, where $\phi$ denotes the trainable parameters of the MLP.
The regularization term $\mathcal{L}_{RAR}$ is computed by aggregating patch-wise similarities across all video frames:
\newcommand{\yhatp}{y_\phi(\mathbf{h}_t)^{[p]}}
\begin{equation}
    \resizebox{0.88\linewidth}{!}{$
    \mathcal{L}_{\mathrm{RAR}}=
    -\sum_{n=1}^{N}
    \mathbb{E}_{t,\mathbf{x}_\ast,\epsilon_t}\Bigl[
      \tfrac{1}{P}\sum_{p=1}^{P}
      \operatorname{sim}\!\bigl(\mathbf{y}_\ast^{[p]},\yhatp\bigr)
    \Bigr].
    $}
\end{equation}
During inference, the external encoder $\mathcal{E}_{d}$ and MLP projector $y_\phi$ are discarded.

\section{Experiments}
\label{sec:exp}

\subsection{Experimental Setup}
\label{sec:setup}

\noindent\textbf{Training Details.} We leverage a small collection (i.e., 150) of high-quality in-house videos as our training data. These video samples encompass various motion (e.g., people running, airplanes taxiing) and appearance changes (e.g., a man transitioning from youth to old age).
The learnable parameters of the backward U-Net are initialized to zero, while the other available weights are initialized from the pre-trained checkpoint of DynamiCrafter \cite{xing2023dynamicrafter}.
We utilize the AdamW optimizer with learning rate of $1 \times e^{-5}$. VTG is trained on 4 NVIDIA A100-80GB GPUs, taking $\sim20K$ iterations with batch size of 2. \\

\noindent\textbf{Inference Details.} In our experiments, we use Stable Diffusion v2.1-base \cite{rombach2022high} as our interpolation backbone.
When training LoRAs, we set the LoRA rank to 16 and train them for 200 steps using AdamW optimizer \cite{loshchilov2018fixing} with a learning rate of $2\times10^{-4}$. This requires only $\sim$ 85s with 1 NVIDIA A100-80GB GPU.
During inversion and sampling, we use the DDIM sampler \cite{song2020denoising} and set total time steps to the default value (i.e., 50) of the selected image-to-video models.
we apply text classifier-free guidance \cite{ho2022classifier} for all models with the same negative prompt ``Distorted, discontinuous, Ugly, blurry, low resolution, motionless, static, disfigured, disconnected limbs, Ugly faces, incomplete arms'' across all transition generations. \\

\noindent\textbf{Comparison Methods.} 
We compare VTG with following methods: DiffMorpher \cite{zhang2023diffmorpher}, TVG \cite{zhang2024tvg}, SEINE \cite{chen2023seine}, DynamiCrafter \cite{xing2023dynamicrafter}, Generative Inbetweening \cite{wang2024generative}, Text Embedding Interpolation (TEI) \cite{wang2023interpolating}, Denoising Interpolation (DI) \cite{he2024aid}, and Attention Interpolation (AID) \cite{he2024aid}. \\

\noindent\textbf{Evaluation Benchmarks.} We employ several public datasets and our self-curated TransitBench to comprehensively evaluate the quality of the generated transition videos.
Specifically, we leverage MorphBench \cite{zhang2023diffmorpher} and TC-Bench \cite{feng2024tc} to evaluate object morphing and motion prediction.
Concept blending and scene transition are rarely explored in the realm of generative models. Hence, there is a lack of specific evaluation benchmarks for these two tasks.
Previously, \cite{he2024aid} employs CIFAR-10 \cite{krizhevsky2009learning} and LAION-Aesthetics \cite{schuhmann2022laion} as their evaluation benchmarks to test concept blending. While there are sufficient number of samples in both datasets, CIFAR-10 only contains concepts from 10 classes, and LAION-Aesthetics consists entirely of synthetic images and lacks real-world samples.
To this end, we present \emph{TransitBench}, the first benchmark dataset for collectively assessing concept blending transitions of two distinct conceptual objects and scene transitions between two relevant scenarios.
We collected 200 pairs of pictures (each pair forms the first and the last frames of one transition generation sample) of diverse content and styles, and evenly divide them into two categories: 1) \emph{concept-blending} cases, and 2) \emph{scene-transition} cases, both of which are obtained from web resources.
We hope \href{https://huggingface.co/datasets/mwxely/TransitBench}{\textcolor{magenta}{TransitBench}} can promote future studies on general transition generation.

\subsection{Qualitative Results}
\label{sec:qual}

\begin{figure*}
    \centering
    \includegraphics[width=0.95\textwidth]{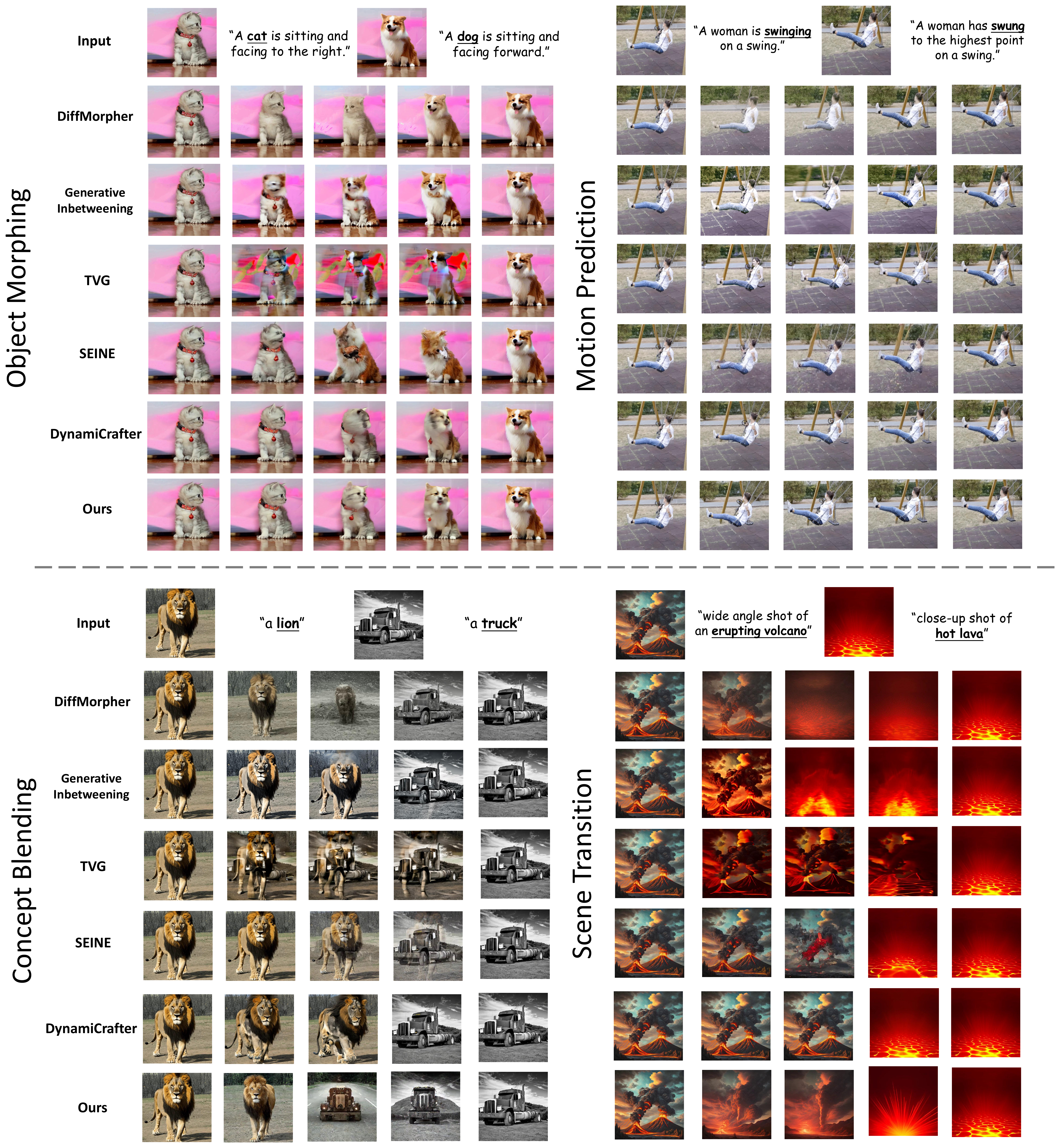}
    \captionsetup{width=\textwidth}
    \caption{\textbf{Qualitative comparisons on four transition tasks.} For each task, the displayed images correspond to the 1st, 4th, 8th, 12th, and 16th video frames, respectively. Please refer to our \href{https://mwxely.github.io/projects/yang2025vtg/index}{\textcolor{magenta}{project page}} for more video examples, which offers a more straightforward and comprehensive qualitative comparisons.}
    \label{fig:quali}
\end{figure*}

To demonstrate the superiority of our method, we provide a visual comparison of VTG against five state-of-the-art baselines.
Specifically, we compare our approach with DiffMorpher, Generative Inbetweening, TVG, SEINE, DynamiCrafter across four transition tasks for uniformity.
\Cref{fig:quali} shows the generation results by different methods.
For object morphing, the compared methods exhibit low-fidelity (e.g., DiffMorpher, DynamiCrafter), oversaturation effect (e.g., TVG), and inaccurate semantics (e.g., Generative Inbetweening, SEINE). In contrast, VTG presents a natural transition, preserving the content of the first and last frames well.
For motion prediction, Generative Inbetweening and SEINE generate inferior transitions as observed by their blurry intermediate frames. TVG yields unnatural transitions as it generates a front-facing image of a person, while the person's face is actually turned away. DiffMorpher shows low-fidelity transitions, and its generated video looks like a sequence of intermittent images without concerning the consistency between background and foreground objects. This is largely due to the missing temporal modeling in DiffMorpher's image diffusion backbone. VTG achieves comparable or even better results with DynamiCrafter, validating that it can effectively preserve the appearance and motion priors in diffusion models.
For concept blending, most video diffusion baselines present abrupt content changes, while other baselines show blurry (e.g., DiffMorpher) and ambiguous (e.g., TVG) intermediate frames. In contrast, VTG yields smooth transition between the lion and the truck with rational intermediate results (e.g., a lion-colored and lion-sized truck).
For scene transition between two relevant scene images, existing methods suffer from abrupt content changes or low similarity (e.g., TVG), while VTG produces natural scene-level transition with pleasing effect.

\subsection{Quantitative Results}
\label{sec:quan}

\noindent\textbf{Evaluation Metrics.}
The objective of \emph{versatile transition generation} is to generate a sequence of $(N-2)$ intermediate video frames, denoted as $x^{2:N-1}$, between two source frames $x^{1}$ and $x^{N}$, with two captions describing the two frames.
High-quality transition videos should simultaneously satisfy four criteria: semantic fidelity to the input endpoints, visual quality, temporal coherence, and perceptual smoothness.
To ensure fair comparisons, we performed extensive quantitative evaluations using experimental settings consistent with those prior methods as in their original paper.
Specifically, we adopt the following evaluation metrics:
(1) FID ($\downarrow$) and PPL ($\downarrow$) follow DiffMorpher \cite{zhang2023diffmorpher}. FID compares the first and last frames against the generated transition frames in the Inception feature space; PPL measures the average LPIPS fluctuation along an interpolation path, where lower values imply fewer perceptual jumps.
(2) TCR ($\uparrow$) and TC-Score ($\uparrow$) are taken from the training-free TVG framework \cite{zhang2024tvg}. TCR is the fraction of video frames whose CLIP embedding stays within a pre-set similarity margin to either endpoint; TC-Score complements it with a mean cosine-similarity, offering a softer view of frame-level consistency.
(3) Smoothness ($\uparrow$) is adapted from AID \cite{he2024aid}. For every adjacent frame pair, we compute LPIPS and the Gini coefficient of these distances, reflecting how even the step-to-step changes are. We report its inverse so that a higher score denotes smoother transitions.

\begin{table}[ht!]
    \centering
    \resizebox{\linewidth}{!}{
        \begin{tabular}{lcccc}
            \toprule
            \multirow{2}{*}{\textbf{Method}} & \multicolumn{2}{c}{\textit{\textbf{Metamorphosis}}} &
                     \multicolumn{2}{c}{\textit{\textbf{Animation}}} \\
            \cmidrule(lr){2-3}\cmidrule(lr){4-5}
             & FID (↓) & PPL (↓) & FID (↓) & PPL (↓) \\
            \midrule
            DiffMorpher \cite{zhang2023diffmorpher} & \underline{70.49} & \textbf{18.19} & \underline{43.15} & \textbf{5.14} \\
            TVG \cite{zhang2024tvg} & 86.92 & 35.18 & 42.99 & 12.46 \\
            SEINE \cite{chen2023seine} & 82.03 & 47.72 & 48.25 & 16.26 \\
            DynamiCrafter \cite{xing2023dynamicrafter} & 87.32 & 42.09 & 43.31 & \underline{11.16} \\
            \rowcolor{gray!20}
            VTG (Ours) & \textbf{67.39} & \underline{22.80} & \textbf{39.16} & \textbf{5.14} \\
            \bottomrule
        \end{tabular}
    }
    \caption{\textbf{Quantitative results on MorphBench.} 
             The best results are in \textbf{bold}; second-best are \underline{underlined}.}
    \label{tab:exp_morphbench}
\end{table}

\begin{table}[ht!]
    \centering
    \resizebox{\linewidth}{!}{
        \begin{tabular}{lccccccc}
            \toprule
            \multirow{2}{*}{\textbf{Method}} & \multicolumn{2}{c}{\textit{\textbf{Attribute}}} &
            \multicolumn{2}{c}{\textit{\textbf{Object}}} &
            \multicolumn{2}{c}{\textit{\textbf{Background}}} \\
            \cmidrule(lr){2-3}\cmidrule(lr){4-5}\cmidrule(lr){6-7}
            & TCR (↑) & TC-Score (↑)
            & TCR (↑) & TC-Score (↑)
            & TCR (↑) & TC-Score (↑) \\
            \midrule
            DiffMorpher \cite{zhang2023diffmorpher} & 41.82 & 0.844 & 19.57 & 0.765 & \textbf{50.00} & 0.819 \\
            SEINE \cite{chen2023seine} & 17.86 & 0.720 & 10.48 & 0.654 &  7.96 & 0.742 \\
            DynamiCrafter \cite{xing2023dynamicrafter} & 16.55 & 0.745 & 13.91 & 0.707 & 25.56 & 0.795 \\
            TVG \cite{zhang2024tvg} & 41.82 & 0.877 & 30.44 & 0.822 & 38.89 & 0.864 \\
            \rowcolor{gray!20}
            VTG (Ours) & \textbf{42.78} & \textbf{0.893} &
                              \textbf{33.46} & \textbf{0.849} &
                              \textbf{50.00} & \textbf{0.883} \\
            \bottomrule
        \end{tabular}
    }
    \caption{\textbf{Quantitative results on TC-Bench.} 
             The best results are in \textbf{bold}. Best viewed when zoomed in.}
    \label{tab:exp_tc_i2v}
\end{table}

\begin{table}[ht]
    \centering
    \renewcommand{\arraystretch}{1.1}
    \resizebox{\linewidth}{!}{
    \begin{tabular}{lcccc>{\columncolor{gray!20}}c}
        \toprule
        \textbf{Dataset} & \textbf{TEI} & \textbf{DI} & \textbf{AID-O} & \textbf{AID-I} & \textbf{Ours} \\
        \midrule
        CIFAR-10         & 0.7531 & 0.7564 & 0.7831 & 0.7861 & \textbf{0.7932} \\
        LAION-Aesthetics & 0.7424 & 0.7511 & 0.7643 & 0.8152 & \textbf{0.8215} \\
        \bottomrule
    \end{tabular}}
    \caption{\textbf{Smoothness (\(\uparrow\)) evaluation for Concept Blending.} 
    The best results are in \textbf{bold}.}
    \label{tab:exp_aid}
\end{table}

\noindent\textbf{Quantitative Results.} As seen in \Cref{tab:exp_morphbench} and \Cref{tab:exp_tc_i2v}, VTG outperforms existing methods with their original experimental setting, demonstrating its capability to generate high-quality, temporally coherent, and semantically consistent video transitions.
It is worth noting that the only exception is \textit{Metamorphosis} subset, where VTG achieves the second-best PPL (22.80). Nevertheless, VTG still demonstrates outstanding overall performance, achieving state-of-the-art FID and comparable PPL.

According to AID \cite{he2024aid}, a well-interpolated concept-blending sequence should exhibit a gradual and smooth transition.
To offer a more comprehensive evaluation, we conducted experiments using the same benchmarks and metric as in AID. Please refer to \Cref{tab:exp_aid} for details. \\

\begin{figure}[ht!]
    \centering
    \includegraphics[width=\linewidth]{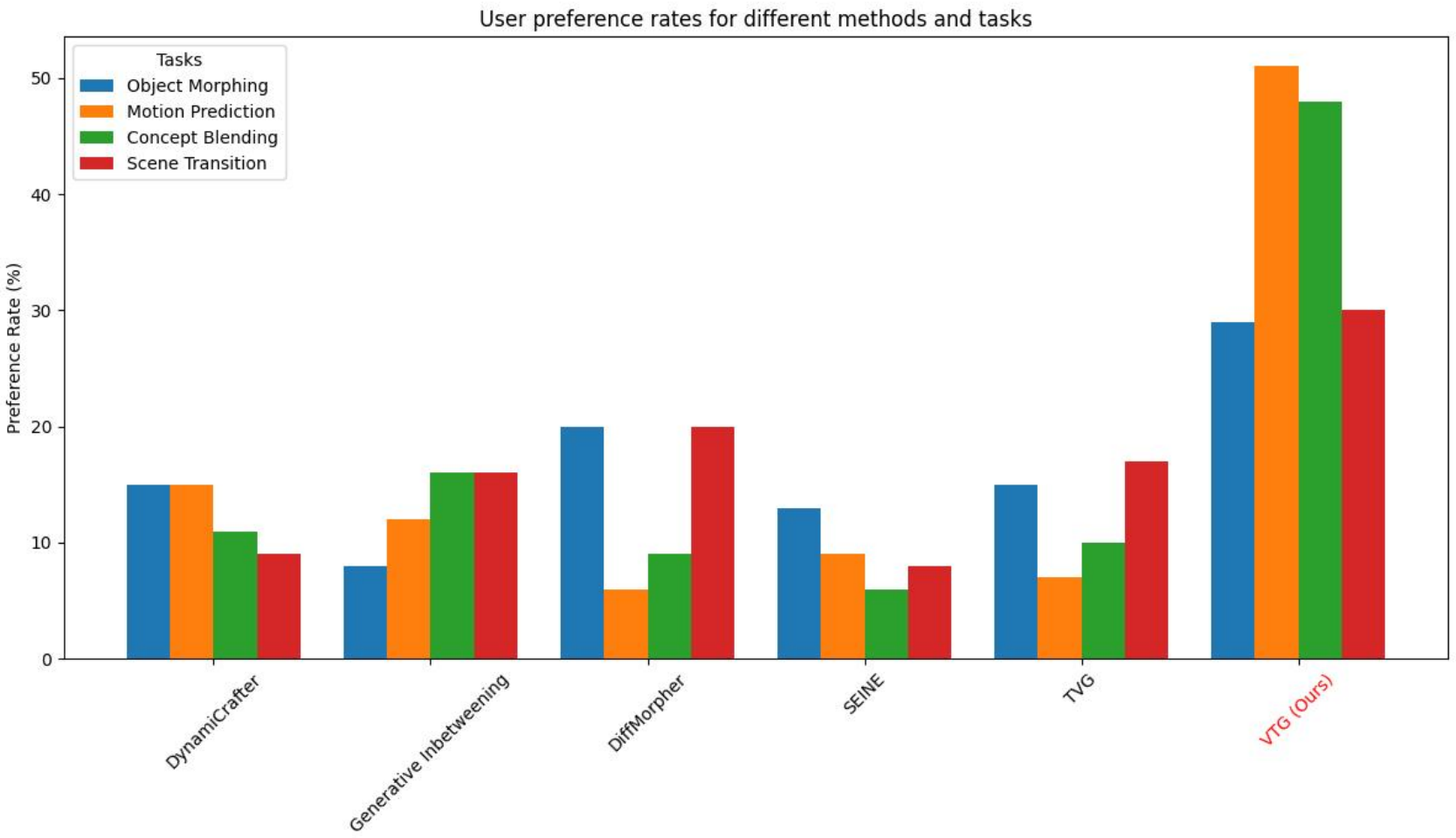}
    \caption{\textbf{The bar plot of user preference rates (\%) of different methods on four transition tasks.} Best viewed when zoomed in.}
    \label{fig:user_pref}
\end{figure}

\noindent\textbf{User Study.} To subjectively evaluate our approach , we further conduct a user study.
We asked Amazon Mechanical Turk (AMT) workers to choose the best generated transition video from a set of five candidates.
Given the two input frames and corresponding transition captions, AMT workers evaluate the videos based on three questions: 
1) \emph{Which video is the most semantic-coherent?}
2) \emph{Which video is the most temporally coherent?}
3) \emph{Which video has the highest fidelity?}
To make a fair evaluation, we include static and obviously incorrect videos to identify random clicking.
Each test example is assigned to 10 different workers for evaluation. 
We collected 150 valid responses for each transition task after filtering out those responses from random clicking workers.
The results are shown in \Cref{fig:user_pref}, which illustrates that VTG obtains the highest preference rate with its superior quality of the generated transition videos compared to other baselines.

\section{Conclusion}
\label{sec:conclusion}

In this paper, we formulate the problem of \emph{versatile transition generation} that encompasses four representative transition tasks, namely, object morphing, motion prediction, concept blending, and scene transition. Leveraging a pretrained image-to-video diffusion model, we design \textbf{VTG}, a unified \underline{\textbf{V}}ersatile \underline{\textbf{T}}ransition video \underline{\textbf{G}}enerator that can tackle the four tasks within a single framework. Both qualitative and quantitative experiments show that VTG achieves superior performance in generating semantically relevant, temporally coherent, and visually pleasing transition video frames, given a pair of text prompts and two input frames.
With these advantages, VTG can be readily applied to real-world content creation scenarios, offering an efficient tool for generating high-quality transition in media production.
\section*{Acknowledgements}
This study is funded by the Ministry of Education Singapore, under the Tier-1 project scheme with the project number RT18/22.

{
    \small
    \bibliographystyle{ieeenat_fullname}
    \bibliography{main}
}

\end{document}